\documentclass[10pt,twocolumn,letterpaper]{article}

\usepackage[pagenumbers]{cvpr} % To force page numbers, e.g. for an arXiv version

\usepackage{graphicx}
\usepackage{amsmath}
\usepackage{amssymb}
\usepackage{booktabs}
\usepackage{url}
\usepackage{caption}
\usepackage{subcaption}
\usepackage{multirow}
\usepackage[accsupp]{axessibility}  % Improves PDF readability for those with disabilities.

\usepackage{xcolor}
\definecolor{RoyalBlue}{rgb}{0,0,0.8}
\usepackage[pagebackref,breaklinks,colorlinks=true,linkcolor=RoyalBlue,citecolor=RoyalBlue]{hyperref}

\usepackage[capitalize]{cleveref}
\crefname{section}{Sec.}{Secs.}
\Crefname{section}{Section}{Sections}
\Crefname{table}{Table}{Tables}
\crefname{table}{Tab.}{Tabs.}

\def\cvprPaperID{10272} % *** Enter the CVPR Paper ID here
\def\confName{CVPR}
\def\confYear{2022}

\begin{document}

%%%%%%%%% TITLE - PLEASE UPDATE
\title{Patch-level Representation Learning for Self-supervised Vision Transformers}

\author{Sukmin Yun\quad\quad Hankook Lee\quad\quad Jaehyung Kim\quad\quad Jinwoo Shin\\
Korea Advanced Institute of Science and Technology (KAIST)\\
{\tt\small \{sukmin.yun, hankook.lee, jaehyungkim, jinwoos\}@kaist.ac.kr}\\
}

\maketitle

%%%%%%%%% ABSTRACT
\begin{abstract}
Recent self-supervised learning (SSL) methods have shown impressive results in learning visual representations from unlabeled images. This paper aims to improve their performance further by utilizing the architectural advantages of the underlying neural network, as the current state-of-the-art visual pretext tasks for SSL do not enjoy the benefit, {\it i.e.,} they are architecture-agnostic. In particular, we focus on Vision Transformers (ViTs), {which have gained much attention recently as a better architectural choice, often outperforming convolutional networks for various visual tasks. The unique characteristic of ViT is that it takes a sequence of disjoint patches from an image and processes patch-level representations internally.} Inspired by this, we design a simple yet effective visual pretext task, coined \mbox{SelfPatch}, for learning better patch-level representations. To be specific, we enforce invariance against each patch and its neighbors, i.e., each patch treats similar neighboring patches as positive samples. Consequently, training ViTs with SelfPatch learns more semantically meaningful relations among patches (without using human-annotated labels), which can be beneficial, in particular, to downstream tasks of a dense prediction type. Despite its simplicity, we demonstrate that it can significantly improve the performance of existing SSL methods for various visual tasks, including object detection and semantic segmentation. Specifically, SelfPatch significantly improves the recent self-supervised ViT, DINO, by achieving +1.3 AP on COCO object detection, +1.2 AP on COCO instance segmentation, and +2.9 mIoU on ADE20K semantic segmentation.

\end{abstract}

%%%%%%%%% BODY TEXT
\section{Introduction}
\label{sec:intro}
Recently, self-supervised learning (SSL) has achieved successful results in learning visual representations from unlabeled images with a variety of elaborate pretext tasks, including contrastive learning \cite{he2019momentum, chen2020simple, chen2020mocov2}, clustering \cite{caron2020unsupervised}, and pseudo-labeling \cite{grill2020bootstrap,caron2021emerging,chen2021exploring}. The common nature of their designs is on utilizing different augmentations from the same image as the {\it positive pairs}, {\it i.e.}, they learn representations to be invariant to the augmentations. The SSL approaches without utilizing human-annotated labels have been competitive with or even outperformed the standard supervised learning~\cite{he2016deep} in various downstream tasks, including image classification~\cite{chen2020mocov2}, object detection~\cite{caron2020unsupervised}, and segmentation~\cite{caron2020unsupervised}.

Meanwhile, motivated by the success of Transformers \cite{vaswani2017attention} in natural language processing \cite{devlin2018bert,brown2020_gpt3}, Vision Transformers (ViTs) \cite{dosovitskiy2020image, touvron2020training, touvron2021going} have gained much attention as an alternative to convolutional neural networks (CNNs) with superior performance over CNNs in various visual tasks~\cite{touvron2020training, radford2021learning}. For example, ViT-S/16~\cite{touvron2020training} has a $\tt 1.8\times$ faster throughput than ResNet-152~\cite{he2016deep} with higher accuracy in the ImageNet \cite{deng2009imagenet} benchmark.

\begin{figure*}[t]
\centering
\includegraphics[width=0.79\textwidth]{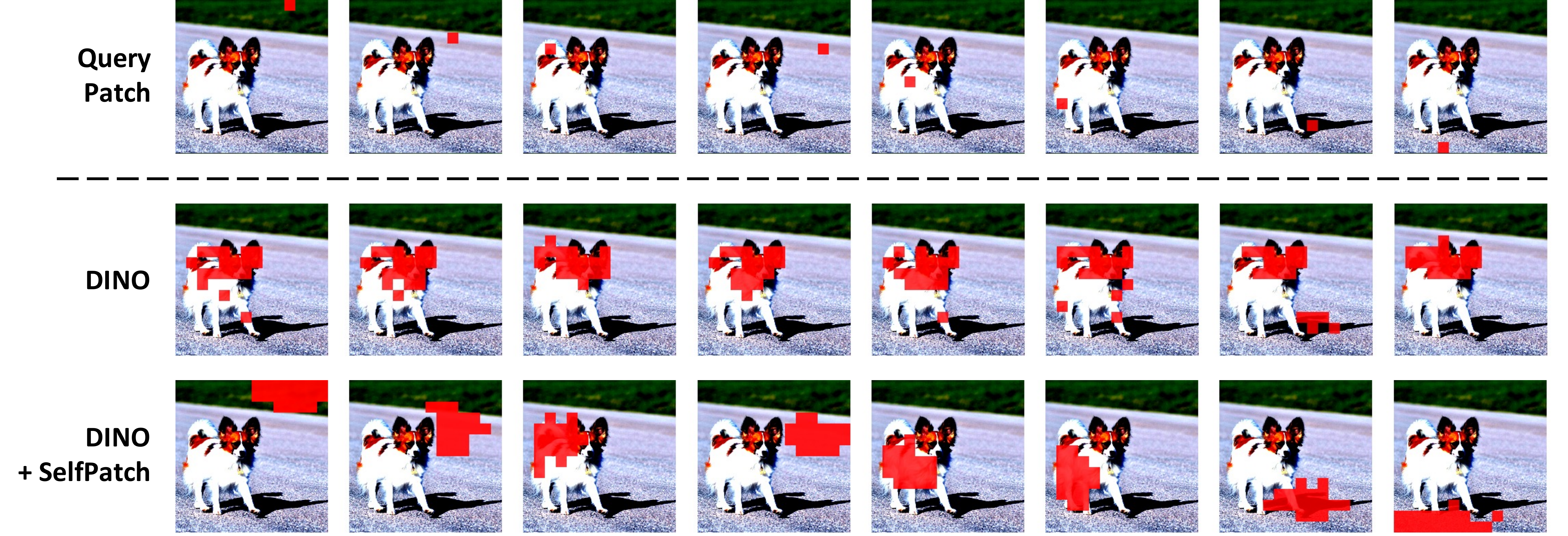}
\vspace{-0.05in}
\caption{Visualization of top-10\% patches obtained by thresholding the self-attention maps of query patches (top) in the last layer of ViT-S/16 trained with DINO (middle) and DINO + SelfPatch (bottom).
While the selected patches obtained by DINO are not semantically correlated with its query patch, SelfPatch encourages the model to learn semantic correlations among patches.
}\label{fig:1}
\vspace{-0.1in}
\end{figure*}

There have been several recent attempts to apply existing self-supervision techniques to ViTs \cite{chen2021empirical, xie2021self, caron2021emerging}. Although the techniques have shown to be also effective with ViTs, they
do not fully utilize the architectural advantages of ViTs, {\it i.e.}, their pretext tasks are architecture-agnostic. For example, ViTs are able to process patch-level representations, but pretext tasks used in the existing SSL schemes only use the whole image-level self-supervision
without considering learning patch-level representations.
As a result, existing self-supervised ViTs~\cite{chen2021empirical, caron2021emerging} may fail to capture semantically meaningful relations among patches; \eg, collapsed self-attention maps of among patches as shown in the second row of \cref{fig:1}.
This limitation inspires us to investigate the following question: {\it how to utilize architectural characteristics of ViTs for improving the quality of learned patch-level representations without human-annotated supervision?}

\vspace{0.03in}
\noindent{\bf Contribution.}
In this paper, 
{we propose a simple yet effective SSL scheme for learning patch-level representations, coined \emph{SelfPatch},} 
which can be beneficial to various visual downstream tasks. 
Our patch-level SSL scheme, SelfPatch, can be incorporated into any image-level self-supervised ViT, \emph{e.g.,} DINO \cite{caron2021emerging}, MoCo-v3~\cite{chen2021empirical}
and MoBY~\cite{xie2021self}, for learning both global ({\it i.e.}, image-level) and local ({\it i.e.}, patch-level) information simultaneously. \cref{fig:1} shows that SelfPatch enhances the quality of self-attention maps of DINO, which is evidence that SelfPatch encourages to learn better patch-level representations.

Our key idea is to treat \emph{semantically similar neighboring} patches as positive samples motivated by the following prior knowledge: adjacent patches often share a common semantic context. Since there might be multiple positive patches (but we do not know exactly which patches are positive), we first select a fixed number of adjacent patches for \emph{positive} candidates using the cosine similarity between patch representations of the current model. Here, some of them might still be noisy (\eg, not positive), and for the purpose of denoising, we summarize their patch representations by utilizing an additional attention-based aggregation module on the top of ViT.
Then, we minimize the distance between each patch representation and the corresponding summarized one.
We provide an overall illustration of the proposed scheme in \cref{fig:method}.

To demonstrate the effectiveness of our method, we pre-train ViT-S/16~\cite{touvron2020training} on the ImageNet \cite{deng2009imagenet} dataset and evaluate transferring performance on a wide range of dense prediction downstream tasks: 
(a) COCO object detection and instance segmentation~\cite{lin2014microsoft},
(b) ADE20K semantic segmentation~\cite{zhou2017scene}, 
and (c) DAVIS 2017 video object segmentation~\cite{pont20172017}.
Specifically, our method improves the state-of-the-art image-level self-supervision, DINO~\cite{caron2021emerging}, with a large margin, \eg, +1.3 AP$^{\text{bb}}$ (\ie, 40.8 $\rightarrow$ 42.1) on COCO detection (see \cref{tbl:coco}), and +2.9 mIoU (\ie, 38.3 $\rightarrow$ 41.2) on ADE20K segmentation (see \cref{tbl:ade}). As a result, our method outperforms all the SSL baselines~\cite{chen2020mocov2,caron2020unsupervised,wang2021dense,xie2021detco,xiao2021region,xie2021self} such as DenseCL \cite{wang2021dense}. Furthermore, 
we demonstrate high compatibility of SelfPatch by incorporating
with various objectives (\eg, MoBY \cite{xie2021self}), architectures (\eg, Swin Transformer \cite{liu2021swin}), and patch sizes (\eg, $8\times8$). For example, SelfPatch improves MoBY \cite{xie2021self} for both ViT and Swin Transformer by +4.8 and +5.6 $(\mathcal{J}\&\mathcal{F})_m$, respectively, on the DAVIS video segmentation benchmark \cite{pont20172017} (see \cref{table:moby}).

Overall, our work highlights the importance of learning patch-level representations during pre-training ViTs in a self-supervised manner. We hope that ours could inspire researchers to rethink the under-explored problem and provide a new direction of patch-level self-supervised \mbox{learning.}

\section{Method}
\begin{figure*}
  \centering
  \hspace{0.2in}
  \begin{subfigure}[b]{0.68\linewidth}
    \includegraphics[width=\textwidth]{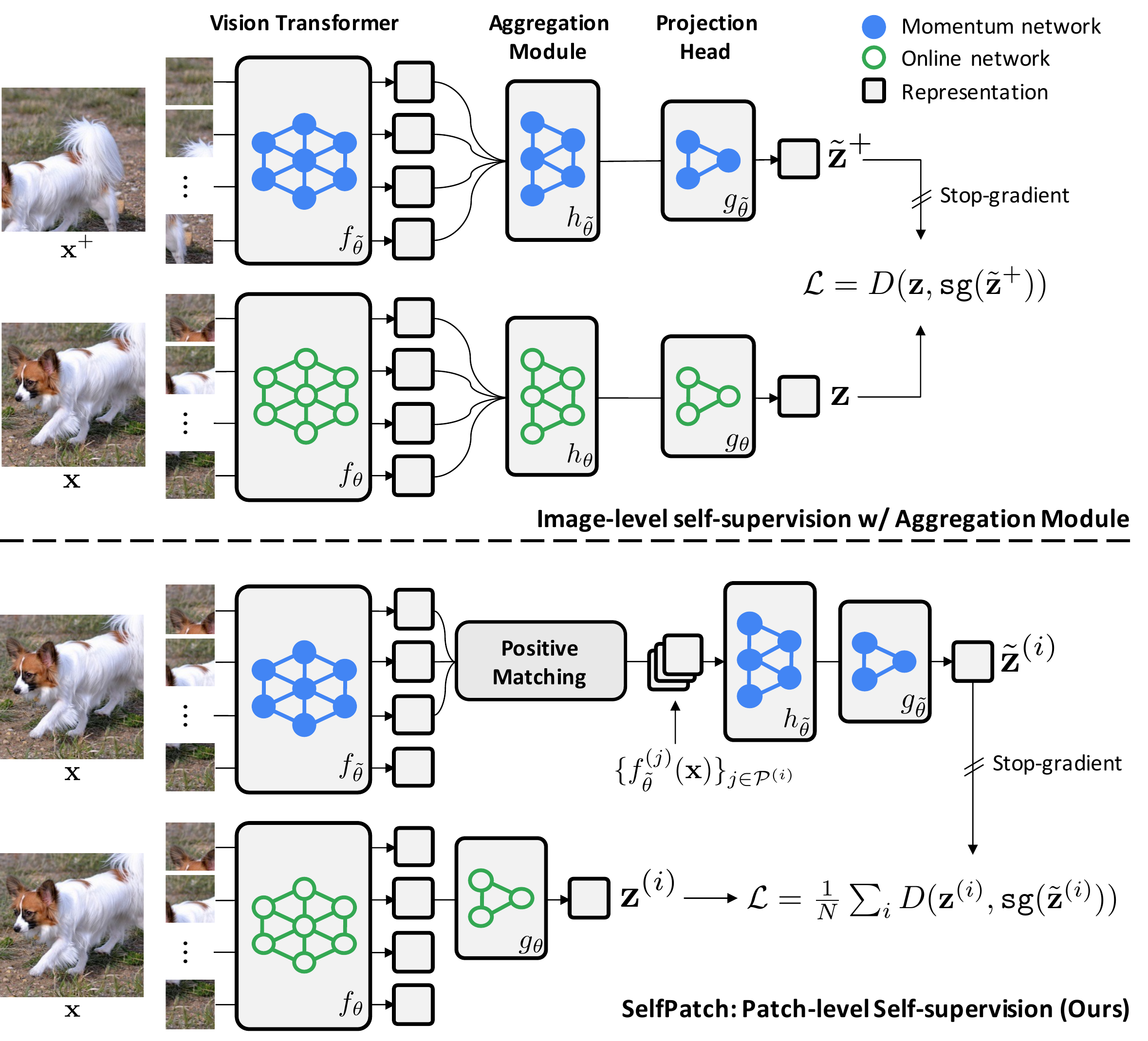}
    \caption{Illustration of the proposed patch-level self-supervision (SelfPatch)}
    \label{fig:method}
  \end{subfigure}
  \hfill
  \begin{subfigure}[b]{0.21\linewidth}
    \includegraphics[width=\textwidth]{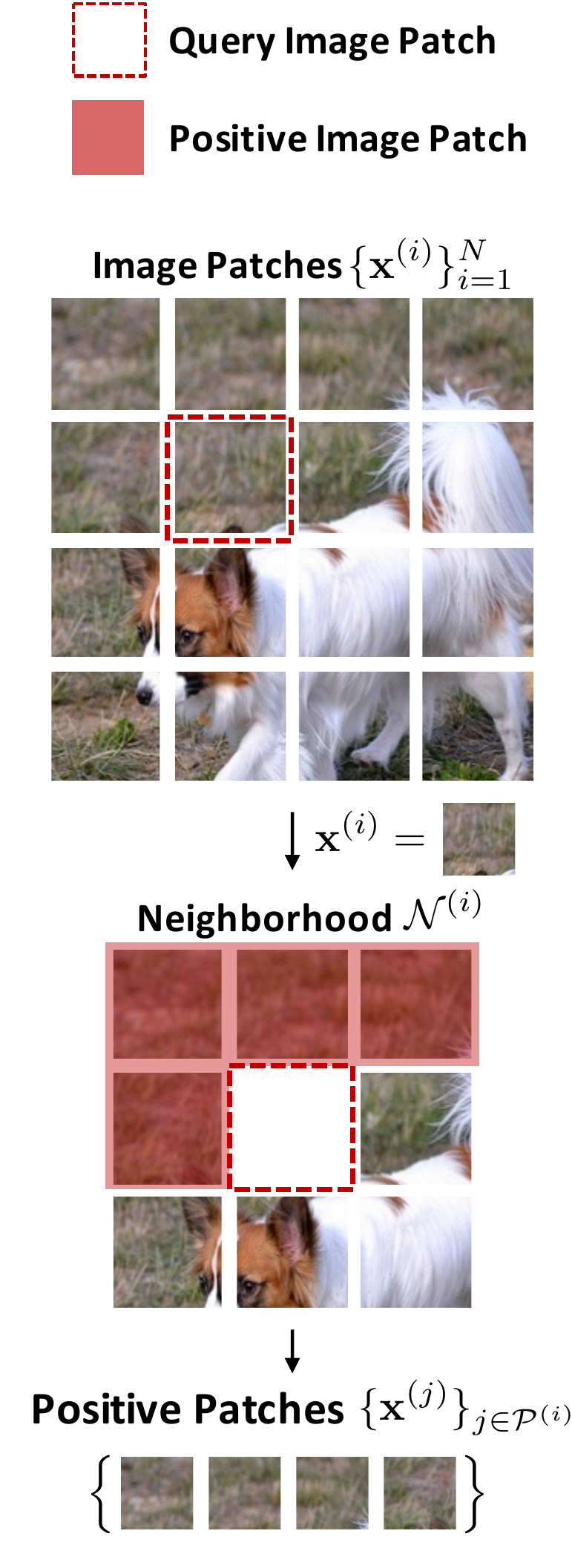}
    \caption{Positive matching}
    \label{fig:matching}
  \end{subfigure}
  \hspace{0.2in}
  \caption{
  (a) \textbf{Top}: image-level self-supervision, which minimizes the distance between the final representations of two differently augmented images. \textbf{Bottom}: patch-level self-supervision, 
  SelfPatch, 
  which minimizes the distance between the final representations of each patch and its positives. 
  We use both types of self-supervision for learning image-level and patch-level representations simultaneously.
  (b) For a given query patch, we find semantically similar
  patches from its neighborhood using
  the cosine similarity on the representation space.}
  \label{fig:overview}
\end{figure*}

In this section, we introduce a simple yet effective visual pretext task,
coined \emph{SelfPatch},
for learning better patch-level representations, which is tailored to Vision Transformers \cite{dosovitskiy2020image} for utilizing their unique architectural advantages. 
We first review Vision Transformers with recent self-supervised learning schemes~\cite{chen2021empirical,caron2021emerging,xie2021self} in \cref{sec:method:pre} and then present details of SelfPatch in \cref{sec:method:our}. 
\cref{fig:method} illustrates the overall scheme of our method, SelfPatch.

\subsection{Preliminaries}\label{sec:method:pre}
\noindent{\bf Vision Transformers.} 
Let $\mathbf{x}\in\mathbb{R}^{H\times W \times C}$ be an image where $(H,W)$ is the resolution of $\mathbf{x}$ and $C$ is the number of channels. Vision Transformers (ViTs) \cite{dosovitskiy2020image} treat the image $\mathbf{x}$ as a sequence of non-overlapping patches $\{\mathbf{x}^{(i)}\in\mathbb{R}^{P^2C}\}_{i=1}^N$ ({\it i.e.}, tokens) where each patch has a fixed resolution $(P, P)$. Then, the patches are linearly transformed to $D$-dimensional patch embeddings $\mathbf{e}^{(i)}=E\mathbf{x}^{(i)}+E_{\text{pos}}^{(i)}\in\mathbb{R}^D$ where $E\in\mathbb{R}^{D\times P^2 C}$ is a linear projection and $E_{\text{pos}}^{(i)}\in\mathbb{R}^{D}$ is a positional embedding for the patch index $i$. ViTs also prepend the $\mathtt{[CLS]}$ token, 
which represents the entire patches ({\it i.e.,} the given image $\mathbf{x}$),
to the patch sequence with a learnable embedding 
$\mathbf{e}^\mathtt{[CLS]}\in\mathbb{R}^D$. 
The resulting input sequence $\mathbf{e}$ is $\mathbf{e}=[\mathbf{e}^\mathtt{[CLS]};\mathbf{e}^{(1)};\mathbf{e}^{(2)};\ldots;\mathbf{e}^{(N)}]$. 
Then, ViTs take the input $\mathbf{e}$ and output all the patch-level and image-level ({\it i.e.}, $\mathtt{[CLS]}$ token) representations with a transformer encoder.\footnote{We omit the details of the transformer encoder \cite{vaswani2017attention} of which each layer consists of the self-attention module, skip connection, and multi-layer perceptron (MLP).} 
For conciseness, we use $f_\theta$ to denote the whole process of a ViT parameterized by $\theta$\footnote{Note that $\theta$ contains all the transformer encoder parameters and embedding parameters $E$, $E_\text{pos}$, and $\mathbf{e}^{\mathtt{[CLS]}}$.} as follows:
\begin{align}
f_\theta(\mathbf{x})
&=f_\theta([\mathbf{e}^\mathtt{[CLS]};\mathbf{e}^{(1)};\mathbf{e}^{(2)};\ldots;\mathbf{e}^{(N)}])\nonumber\\
&=[f_\theta^\mathtt{[CLS]}(\mathbf{x});f_\theta^{(1)}(\mathbf{x});f_\theta^{(2)}(\mathbf{x});\ldots;f_\theta^{(N)}(\mathbf{x})],
\end{align}
where $f_\theta^\mathtt{[CLS]}(\mathbf{x})$ and $f_\theta^{(i)}(\mathbf{x})$ are the final representations of the $\mathtt{[CLS]}$ token and the $i$-th patch, respectively. Remark that $f_\theta^\mathtt{[CLS]}(\mathbf{x})$ is utilized for solving image-level downstream tasks such as image classification \cite{dosovitskiy2020image,touvron2020training,touvron2021going} 
while the patch-level representations $\{f_\theta^{(i)}(\mathbf{x})\}_{i=1}^N$ are done for dense prediction tasks, \emph{e.g.,} object detection~\cite{carion2020end} and semantic segmentation~\cite{xie2021segmenting}. 

\vspace{0.03in}
\noindent{\bf Self-supervised learning with ViTs.}
The recent literature \cite{chen2021empirical, xie2021self, caron2021emerging} has attempted to apply self-supervised learning (SSL) schemes to ViTs. They commonly construct a positive pair $(\mathbf{x},\mathbf{x}^+)$ by applying different augmentations to the same image, and then learn their representations to be similar, {\it i.e.}, invariant to augmentations. We here provide a generic formulation of this idea.\footnote{Built upon this formulation, one can additionally consider negative pairs for contrastive learning or asymmetric architectures such as a prediction head \cite{chen2021empirical,xie2021self}.} 
To this end, we denote two ViT backbone networks as $f_\theta$ and $f_{\tilde\theta}$, and their projection heads as $g_\theta$ and $g_{\tilde\theta}$ are parametrized by $\theta$ and $\tilde\theta$, respectively. 
Then, the generic form of \emph{image-level} self-supervision $\mathcal{L}^{\text{SSL}}(\{\mathbf{x}, \mathbf{x}^{+}\}; \theta, \tilde{\theta})$ can be written as follows:
\begin{equation}
     \mathcal{L}^{\text{SSL}}
     := 
     D(g_{\theta}(f_\theta^\mathtt{[CLS]}(\mathbf{x})),
     {\mathtt{sg}}(g_{\tilde\theta}(f_{\tilde\theta}^\mathtt{[CLS]}(\mathbf{x}^{+})))),\label{loss:SSL}
\end{equation}
where $D$ is a distance function and ${\mathtt{sg}}$ is the stop-gradient operation. 
Note that the choices of distance $D$ and architecture $g$ depend on a type of self-supervision;
for example, Caron \etal~\cite{caron2021emerging} update $\tilde{\theta}$ by the exponential moving average of $\theta$, and use Kullback-Leibler (KL) divergence as $D$, where the projection heads $g_{\tilde{\theta}}$ and $g_{\theta}$ are designed to produce a probability distribution over the final feature dimension. 

We remark that the idea of constructing a positive pair $(\mathbf{x},\mathbf{x}^+)$ 
of 
augmented {\it images} is architecture-agnostic, which means it does not fully utilize the architectural benefits of ViTs. For example, ViTs can handle patch-level representations $\{f_{\theta}^{(i)}(\mathbf{x})\}$, but the recent SSL schemes use only $f_\theta^{\mathtt{[CLS]}}(\mathbf{x})$ as described in \cref{loss:SSL}. This motivates us to explore the following question: {\it how to construct patch-level self-supervision, i.e., positive pairs of patches?}

\subsection{Patch-level self-supervision}\label{sec:method:our}
Recall that our goal is to learn better patch-level representations, which can be beneficial to various 
downstream tasks.
Our key idea is to consider neighboring patches as positive samples based on the continuous nature of image patches. 
Specifically, SelfPatch aims to learn
patch-level
representations via predicting self-supervision constructed by
the following procedure: 
for each patch $\mathbf{x}^{(i)}$, 
\emph{Positive Matching} first finds a set of candidates for positive patch indices 
$\mathcal{P}^{(i)}$ from its \emph{neighborhood} $\mathcal{N}^{(i)}$ (see \cref{fig:matching}), and then \emph{Aggregation Module} aggregates their representations $\{f_\theta^{(j)}(\mathbf{x})\}_{j\in\mathcal{P}^{(i)}}$ as self-supervision 
for 
$\mathbf{x}^{(i)}$ (see \cref{fig:method}).

\vspace{0.03in}
\noindent{\bf Neighboring patches.}
Given
a query patch $\mathbf{x}^{(i)}$, we assume that  
there exists a semantically similar patch $\mathbf{x}^{(j)}$ in its neighborhood $\mathcal{N}^{(i)}$, because neighboring patches 
$\{\mathbf{x}^{(j)}\}_{j\in\mathcal{N}^{(i)}}$ often share a semantic context with the query $\mathbf{x}^{(i)}$.
Let $\{\mathbf{x}^{(j)}\}_{j\in\mathcal{N}^{(i)}}$ be a set of neighboring patches, where $\mathcal{N}^{(i)}$ is a set of patch indices in the neighborhood.
We simply use $\mathcal{N}^{(i)}$ to be adjacent patches (\emph{i.e.,} $|\mathcal{N}^{(i)}|=8$), and empirically found that this choice is important for selecting candidates for positive patches (see \cref{sec:component}
for analysis on the importance of neighboring patches).

\begin{table*}[t]
\centering
\scalebox{0.95}{
% \resizebox{\textwidth}{!}{
\begin{tabular}{lcccccccccc}
\toprule
& & & & \multicolumn{3}{c}{Detection} & \multicolumn{3}{c}{Segmentation} \\
\cmidrule(lr){5-7}\cmidrule(lr){8-10}
Method & Backbone & Epoch & Param.(M) & AP$^{\text{bb}}$ & AP$_{50}^{\text{bb}}$ & AP$_{75}^{\text{bb}}$ & AP$^{\text{mk}}$ & AP$_{50}^{\text{mk}}$ & AP$_{75}^{\text{mk}}$ \\
\midrule
MoCo-v2 \cite{chen2020mocov2}& ResNet50 & 200 & 26 & 38.9 & 59.2 & 42.4 & 35.5 & 56.2 & 37.8\\
SwAV \cite{caron2020unsupervised}& ResNet50 & 200 & 26 & 38.5 & 60.4 & 41.4 & 35.4 & 57.0 & 37.7\\
DenseCL \cite{wang2021dense}& ResNet50 & 200 & 26 & 40.3 & 59.9 & 44.3 & 36.4 & 57.0 & 39.2\\
ReSim \cite{xiao2021region}& ResNet50 & 200 & 26 & 40.3 & 60.6 & 44.2 & 36.4 & 57.5 & 38.9\\
DetCo \cite{xie2021detco}& ResNet50 & 200 & 26 & 40.1 & 61.0 & 43.9 & 36.4 & 58.0 & 38.9\\
\cmidrule(lr){1-10}
MoCo-v3 \cite{chen2021empirical} & ViT-S/16 & 300 & 22 & 39.8 & 62.6 & 43.1 & 37.1 & 59.6 & 39.2\\
MoBY \cite{xie2021self} & ViT-S/16 & 300 & 22 & 41.1 & 63.7 & 44.8 & 37.6 & 60.3 & 39.8\\
\cmidrule(lr){1-10}
DINO \cite{caron2021emerging}& ViT-S/16 & 300 & 22 & 40.8 & 63.4 & 44.2 & 37.3 & 59.9 & 39.5\\
+ SelfPatch (ours) & ViT-S/16 & 200 & 22 & {\textbf{42.1}} & {\textbf{64.9}} & {\textbf{46.1}}
& {\textbf{38.5}} & {\textbf{61.3}} & {\textbf{40.8}}\\
\bottomrule
\end{tabular}
}
\vspace{-0.07in}
\caption{\textbf{COCO object detection and instance segmentation} performances of the recent self-supervised approaches pre-trained on ImageNet.
The metrics AP$^{\text{bb}}$ and AP$^{\text{mk}}$ denote bounding box and mask average precision (AP), respectively. 
}\label{tbl:coco}
\vspace{-0.09in}
\end{table*}

\vspace{0.03in}
\noindent{\bf Matching positives from the neighborhood.}
To sample positive ({\it i.e.}, semantically similar) patches from the neighborhood $\mathcal{N}^{(i)}$, we measure the semantic closeness between the query patch $\mathbf{x}^{(i)}$ and its neighboring patch $\mathbf{x}_\theta^{(j)}$ for all $j\in\mathcal{N}^{(i)}$. To this end, we use the cosine similarity on the representation space, {\it i.e.}, 
\begin{align}
s(i,j)=f_\theta^{(i)}(\mathbf{x})^\top f_\theta^{(j)}(\mathbf{x})/||f_\theta^{(i)}(\mathbf{x})||_2||f_\theta^{(j)}(\mathbf{x})||_2.
\end{align}
We take \emph{top-k} positive patches $\{\mathbf{x}^{(j)}\}_{j\in\mathcal{P}^{(i)}}$ based on the similarity scores $s(i,j)$, where $\mathcal{P}^{(i)}$ is a set of patch indices of \emph{top-k} patches in $\mathcal{N}^{(i)}$. We use $k=|\mathcal{P}^{(i)}|=4$ in our experiments (see \cref{sec:component} for analysis on the effect of $k$). 

\vspace{0.03in}
\noindent{\bf Aggregation module.} 
We aggregate positive patches $\{\mathbf{x}^{(j)}\}_{j\in\mathcal{P}^{(i)}}$ using an aggregation module $h_\theta$ to construct patch-level self-supervision $\mathbf{y}^{(i)}_\theta$ for each patch $\mathbf{x}^{(i)}$. 
We utilize $h_\theta$ for not only patch representations ({\it i.e.}, $\mathbf{y}^{(i)}_\theta$), but also the image-level representation ({\it i.e.}, $\mathbf{y}_\theta$). Formally, 
\begin{align}
    \mathbf{y}_\theta&:=h_\theta(\{f_{\theta}^{(j)}(\mathbf{x})\}_{j=1}^N), \label{eq:global_output}\\
    \mathbf{y}^{(i)}_\theta&:=h_\theta(\{f_{\theta}^{(j)}(\mathbf{x})\}_{j\in\mathcal{P}^{(i)}}).
\end{align}
We follow Touvron \etal~\cite{touvron2021going} to implement the aggregation module $h_\theta$. To be specific, $h_\theta$ is a small Transformer that takes a sequence of representations as an input and outputs a representation of the $\mathtt{[CLS]}$ token.
We found that the aggregation module $h_\theta$ is crucial for generating better self-supervision (see \cref{sec:component} for ablation experiments).

\vspace{0.03in}
\noindent{\bf Training objective.} 
Our training objective $\mathcal{L}^{\text{total}}(\mathbf{x},\mathbf{x}^+)$ consists of the \emph{image-level} and \emph{patch-level} self-supervision objectives, $\mathcal{L}^{\text{SSL}}(\mathbf{x},\mathbf{x}^+)$ and $\mathcal{L}^{\text{SelfPatch}}(\mathbf{x})$, as follows:
\begin{align}
&\mathcal{L}^{\text{SSL}}(\mathbf{x},\mathbf{x}^+):= D(g_{\theta}(\mathbf{y}_\theta),{\mathtt{sg}}(g_{\tilde\theta}(\mathbf{y}_{\tilde{\theta}}))), \\
&\mathcal{L}^{\text{SelfPatch}}(\mathbf{x}):= \frac{1}{N}\sum_{i=1}^N D(g'_{\theta}(f_\theta^{(i)}(\mathbf{x})),{\mathtt{sg}}(g'_{\tilde\theta}({\mathbf{y}}^{(i)}_{\tilde{\theta}}))), \label{loss:PASS}\\
&\mathcal{L}^{\text{total}}(\mathbf{x},\mathbf{x}^+):=\mathcal{L}^{\text{SSL}}(\mathbf{x},\mathbf{x}^+)+\lambda\mathcal{L}^{\text{SelfPatch}}(\mathbf{x}),\label{loss:total}
\end{align}
where $D$ is a distance function, ${\mathtt{sg}}$ is the stop-gradient operation, $\lambda$ is a hyperparameter, $g$ and $g'$ are projection heads for $\mathcal{L}^\text{SSL}$ and $\mathcal{L}^\text{SelfPatch}$, respectively.
Here, our patch-level objective $\mathcal{L}^{\text{SelfPatch}}(\mathbf{x})$ enforces a query patch $\mathbf{x}^{(i)}$ and its positives $\{\mathbf{x}^{(j)}\}_{j\in\mathcal{P}^{(i)}}$ to be similar by minimizing the distance between the patch and aggregated representations, $f_\theta^{(i)}(\mathbf{x})$ and ${\mathbf{y}}^{(i)}_{\tilde{\theta}}$, respectively.
We remark that $\mathcal{L}^{\text{SelfPatch}}(\mathbf{x})$ has an asymmetric form: a query representation $f_\theta^{(i)}(\mathbf{x})$ does not use the aggregation module $h_\theta$ while its (aggregated) target ${\mathbf{y}}^{(i)}_{\tilde{\theta}}$ does. We empirically found that this architectural asymmetry with the stop-gradient operation effectively avoids the mode collapse issue which comes from 
minimizing the distance between patch representations.

\section{Related works}
\noindent{\bf Transformer-based architectures for images.} 
Vision Transformer (ViT) \cite{dosovitskiy2020image, touvron2020training} is a pioneering architecture built on top of Transformer~\cite{vaswani2017attention} for large-scale vision tasks such as ImageNet~\cite{deng2009imagenet} classification by splitting an image into a sequence of patch images. 
Inspired by the success of the Vision Transformers, a number of variants 
\cite{graham2021levit, heo2021rethinking, liu2021swin, pan2021scalable, wang2021pyramid, wu2021cvt, zhang2021multi} 
have been developed. They commonly incorporate convolutional designs into ViTs, {\it e.g.}, a spatial down-sampling operation \cite{pan2021scalable, wu2021cvt} or a hierarchical structure that considers various spatial resolutions \cite{wang2021pyramid, liu2021swin}.

\vspace{0.03in}
\noindent{\bf Self-supervised learning.} 
For learning visual representations from a large number of unlabeled images, self-supervised learning (SSL)~\cite{he2019momentum, chen2020simple, chen2020mocov2, grill2020bootstrap, caron2020unsupervised, caron2021emerging, chen2021empirical, xie2021self} has become a remarkable research direction
as it can be effectively transferred to various downstream tasks like image classification.
The common design of their pretext tasks is to learn visual representations by maximizing the similarity between augmented images originated from the same image.
Recently, there have been several attempts to apply such SSL pretext tasks to ViTs~\cite{xie2021self, chen2021empirical, caron2021emerging}.
Meanwhile, some of studies have focused on to design SSL schemes tailored for dense prediction tasks \cite{pinheiro2020unsupervised, wang2021dense, xie2021detco, xiao2021region}.
For example, ReSim~\cite{xiao2021region} matches overlapping
regions (\ie, sub-images have the same location) between two augmented images, 
while our method matches neighboring regions ({\it i.e.,} image patches) within the same augmented image. 
We remark that utilizing adjacent patches as the positives 
can be viewed as a reasonable way to find meaningful positives without constraints of overlapping regions between augmented images.

\section{Experiments}\label{sec:exp}
{In this section, we evaluate the effectiveness of the proposed self-supervised learning scheme, \emph{SelfPatch},}
through extensive large-scale experiments. Specifically, we compare SelfPatch with existing self-supervised learning (SSL) approaches in various
dense prediction tasks:\footnote{Although our major focus is on dense prediction tasks,
we also evaluate transferring performance to ImageNet linear classification task~\cite{deng2009imagenet}, and observe that our method achieves competitive performance compared to the state-of-the-art methods as presented in the supplementary material.} 
(a) COCO object detection and segmentation \cite{lin2014microsoft} (\cref{exp:coco}), (b) ADE20K segmentation \cite{zhou2017scene} (\cref{exp:ade20k}), and (c) DAVIS video object segmentation \cite{pont20172017} (\cref{exp:davis}). 
More details of experimental setups are described in each section and the supplementary material.

\vspace{0.03in}
\noindent{\bf Baselines.}
We consider a variety of existing SSL methods developed for ResNet~\cite{he2016deep} and ViT~\cite{touvron2020training} architectures:
(a) self-supervised ResNets: MoCo-v2~\cite{chen2020mocov2}, SwAV~\cite{caron2020unsupervised}, DenseCL \cite{wang2021dense}, ReSim~\cite{xiao2021region}, and DetCo~\cite{xie2021detco};
and 
(b) self-supervised ViTs: DINO~\cite{caron2021emerging}, MoCo-v3~\cite{chen2021empirical}, and MoBY~\cite{xie2021self}.
We use ViT-S/16~\cite{touvron2020training} (22M parameters) and ResNet50~\cite{he2016deep} (26M parameters) since they are conventional choices and have the similar number of parameters. 
We denote our method built upon an existing method by “+ SelfPatch”, \emph{e.g.,} DINO + SelfPatch.
We use publicly available ImageNet pre-trained models for the baselines; {the public ResNet models \cite{chen2020mocov2,caron2020unsupervised,wang2021dense,xie2021detco,xiao2021region} and ViT models \cite{caron2021emerging,chen2021empirical,xie2021self} are pre-trained for 200 and 300 epochs, respectively.}

\vspace{0.03in}
\noindent{\bf Implementation details.} 
We incorporate the SelfPatch objective with the state-of-the-art SSL scheme, DINO~\cite{caron2021emerging}, as described in \cref{loss:total}, where we pre-train ViT-S/16 on ImageNet for 200 epochs with a batch size of 1024.
We follow DINO's training details ({\it e.g.}, optimizer, learning rate schedule). 
For the distance function $D$ in \cref{loss:PASS}, we use the KL divergence in $\mathcal{L}^\text{SelfPatch}$ following DINO. 
We use $\lambda=0.1$ unless stated otherwise (see \cref{sec:hyperparameter} and \cref{table:lossweight} for hyperparameter analysis). 
The other training details are provided in the supplementary material.

It is worth noting that our choice of epoch
is due to fair comparisons with ResNet and ViT baselines;
when pre-training ViT-S/16, 
our additional self-supervision increases (around 21\%) training time per epoch
and we consider a smaller number of epoch for DINO + SelfPatch, compared to that of DINO
for a fair comparison.
Nevertheless, we expect that our method can be improved further if trained by 300 epochs as like other public ViT baselines~\cite{caron2021emerging, xie2021self, chen2021empirical}. 

\subsection{COCO object detection and segmentation}\label{exp:coco}
\noindent{\bf Setup.}
We evaluate 
pre-trained models
on the COCO object detection and instance segmentation tasks \cite{lin2014microsoft}.
Here, all models are fine-tuned with Mask R-CNN~\cite{he2017mask} and FPN \cite{lin2017feature} 
under the standard $\tt 1x$ schedule.

\vspace{0.03in}
\noindent{\bf Results.} 
\cref{tbl:coco} shows that our SelfPatch consistently improves DINO in both detection and segmentation tasks, and consequently, DINO + SelfPatch outperforms all the baselines. For example, the bounding box average precision ({\it i.e.}, AP$^{bb}$) of DINO + SelfPatch is $1.3$ points higher than that of DINO.
One can find similar results in the segmentation task; for example, DINO + SelfPatch achieves 38.5 mask average precision ({\it i.e.}, AP$^{mk}$),
which is 1.2 points higher than DINO, and also 2.1 points higher than the best ResNet-based baseline, DenseCL.
We also emphasize that the improvements from DINO + SelfPatch are even achieved with 
a smaller number of pre-training epochs (\emph{i.e.,} 200 epochs) than 
DINO.
These results demonstrate that the advantages of our method are not only high performance, but also training efficiency 
(\eg, DINO + SelfPatch is 1.24× faster than DINO in total running time).

\subsection{ADE20K semantic segmentation}\label{exp:ade20k}
\begin{table}[t]
\centering
\scalebox{0.95}{
% \resizebox{0.95\linewidth}{!}{
\begin{tabular}{lcccc}
\toprule
Method & Backbone & mIoU & aAcc & mAcc \\
\midrule
MoCo-v2 \cite{chen2020mocov2}& ResNet50 & 35.8 & 77.6 & 45.1\\
SwAV \cite{caron2020unsupervised}& ResNet50 & 35.4 & 77.5 & 44.9\\
DenseCL \cite{wang2021dense}& ResNet50 & 37.2 & 78.5 & 47.1\\
ReSim \cite{xiao2021region}& ResNet50 & 36.6 & 78.4 & 46.4\\
DetCo \cite{xie2021detco}& ResNet50 & 37.3 & 78.4 & 46.7\\
% \cmidrule(lr){1-5}
\cmidrule(lr){1-5}
MoCo-v3 \cite{chen2021empirical} & ViT-S/16 & 35.3 & 78.9 & 45.9\\
MoBY \cite{xie2021self} & ViT-S/16 & 39.5 & 79.9 & 50.5\\
\cmidrule(lr){1-5}
DINO \cite{caron2021emerging}& ViT-S/16 & 38.3 & 79.0 & 49.4\\
+ SelfPatch (ours) & ViT-S/16 & \textbf{41.2} & \textbf{80.7} & \textbf{52.1}\\
\bottomrule
\end{tabular}}
\vspace{-0.05in}
\caption{\textbf{ADE20K semantic segmentation} performances of the recent self-supervised approaches pre-trained on ImageNet.
The metrics mIoU, aAcc, and mAcc denote mean intersection of union, all pixel accuracy, and mean class accuracy, respectively.
}\label{tbl:ade}
\vspace{-0.15in}
\end{table}

\noindent{\bf Setup.}
We evaluate semantic segmentation performances of pre-trained models on ADE20K \cite{zhou2017scene}, which contains 150 fine-grained semantic categories and 25k training data. All models are fine-tuned with Semantic FPN~\cite{kirillov2019panoptic} under the standard 40k iteration schedule. We report three metrics: (a) mean intersection of union (mIoU) averaged over all semantic categories, (b) all pixel accuracy (aAcc), and (c) mean class accuracy (mAcc).

\vspace{0.03in}
\noindent{\bf Results.} 
As shown in \cref{tbl:ade}, DINO + SelfPatch achieves significant improvements over DINO in all the metrics; \emph{e.g.,} DINO + SelfPatch achieves 2.9 and 2.7 points higher than DINO in terms of the mIoU and mAcc metrics, respectively. Also, DINO + SelfPatch consistently outperforms all the baselines; \emph{e.g.,} in terms of the mIoU metric, our method achieves 41.2 points, while DetCO and MoBY do 37.3 and 39.5 points, respectively. These comparisons across the architectures verify the effectiveness of SelfPatch.

\subsection{DAVIS video object segmentation}\label{exp:davis}
\noindent{\bf Setup.}
We perform video object segmentation using pre-trained models on DAVIS 2017 \cite{pont20172017}.
We follow the evaluation protocol in Caron \etal \cite{caron2021emerging}, which does not require extra training costs. To be specific, it evaluates the quality of frozen representations of image patches by segmenting scenes with the nearest neighbor between consecutive frames. We report three evaluation metrics: (a) mean region similarity $\mathcal{J}_m$, (b) mean contour-based accuracy $\mathcal{F}_m$, and (c) their average score $(\mathcal{J}\& \mathcal{F})_m$.

\begin{figure*}[t]
\centering
%\framebox[4.0in]{$\;$}
\includegraphics[width=0.84\textwidth]{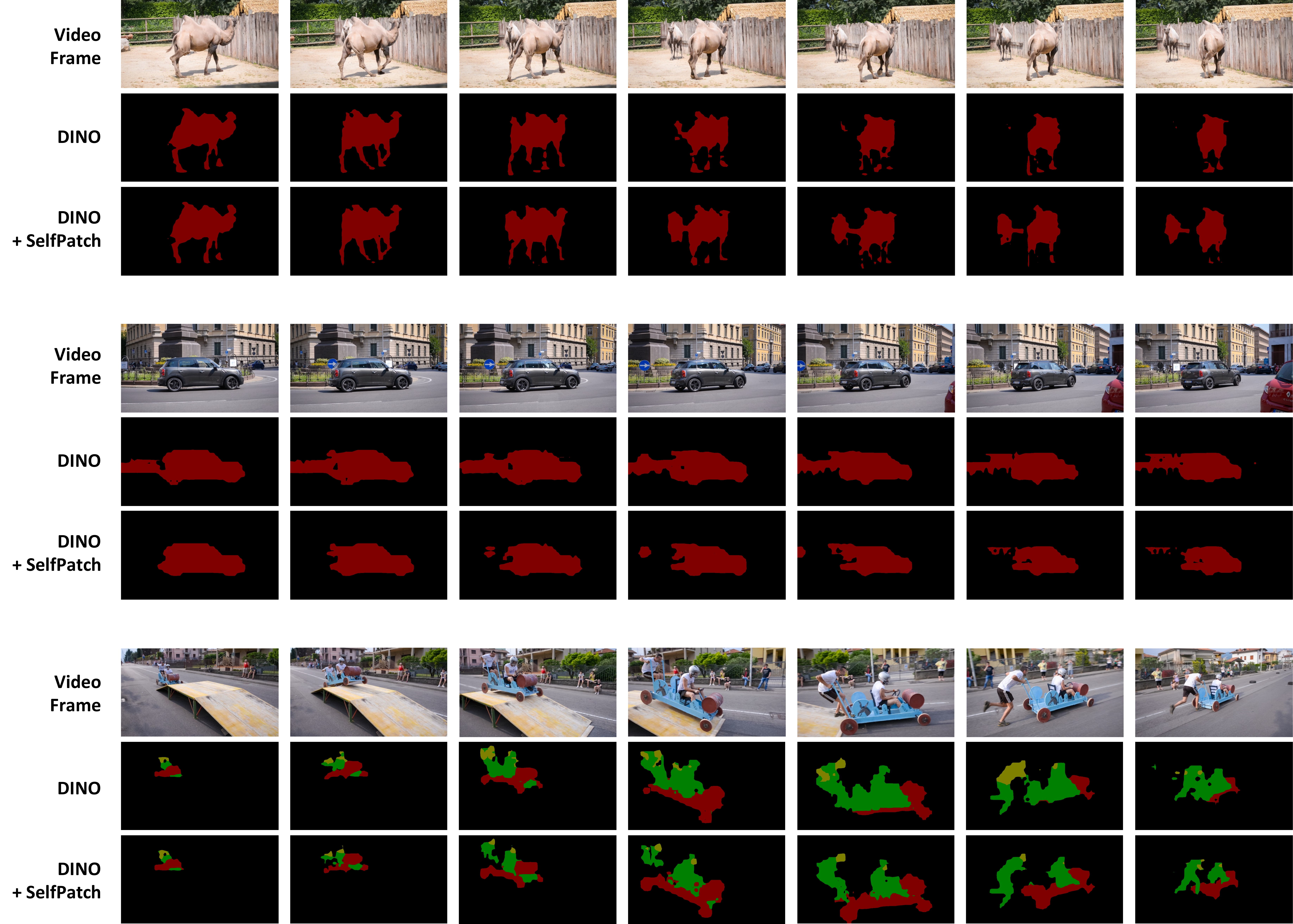}
\vspace{-0.05in}
\caption{Visualization of video segmentation results on the DAVIS 2017~\cite{pont20172017} benchmark. \textbf{Top}: input video frames. \textbf{Middle} and \textbf{Bottom}: segmentation results obtained by DINO and DINO + SelfPatch (ours), respectively. SelfPatch clearly improves the segmentation results, which is evidence that SelfPatch encourages the patch representations to learn semantic information of each object. Best viewed in color.}\label{fig:3}
\vspace{-0.1in}
\end{figure*}

\vspace{0.03in}
\noindent{\bf Results.} 
In \cref{tbl:davis}, DINO + SelfPatch does not only consistently improve DINO, but also largely surpasses the other baselines. For example, SelfPatch improves the $(\mathcal{J}\& \mathcal{F})_m$ score of DINO from 60.7 to 62.7, while SwAV and MoBY achieve only 57.4 and 54.7, respectively.
We present visualizations of video object segmentation results obtained by DINO and DINO + SelfPatch in \cref{fig:3}, and it shows that
SelfPatch clearly enhances the video segmentation quality.
From these observations, we confirm that our method encourages the model to produce more meaningful segmentation maps, which also demonstrate the effectiveness of our scheme for learning better patch-level representations.

\section{Ablation study}\label{sec:ablation}
\begin{table}[t]
\centering
\scalebox{0.95}{
% \resizebox{0.95\linewidth}{!}{
\begin{tabular}{lcccc}
\toprule
Method & Backbone & $(\mathcal{J}\&\mathcal{F})_m$ & $\mathcal{J}_m$ & $\mathcal{F}_m$ \\
\midrule
MoCo-v2 \cite{chen2020mocov2}& ResNet50 & 55.5 & 56.0 & 55.0\\
SwAV \cite{caron2020unsupervised}& ResNet50 & 57.4 & 57.6 & 57.3\\
DenseCL \cite{wang2021dense}& ResNet50 & 50.7 & 52.6 & 48.9\\
ReSim \cite{xiao2021region}& ResNet50 & 49.3 & 51.2 & 47.3\\
DetCo \cite{xie2021detco}& ResNet50 & 56.7 & 57.0 & 56.4\\
\cmidrule(lr){1-5}
MoCo-v3 \cite{chen2021empirical} & ViT-S/16 & 53.5 & 51.2 & 55.9\\
MoBY \cite{xie2021self} & ViT-S/16 & 54.7 & 52.0 & 57.3\\
\cmidrule(lr){1-5}
DINO \cite{caron2021emerging}& ViT-S/16 & 60.7 & 59.1 & 62.4\\
+ SelfPatch (ours) & ViT-S/16 & {\textbf{62.7}} & {\textbf{60.7}} & {\textbf{64.7}}\\
\bottomrule
\end{tabular}}
\vspace{-0.05in}
\caption{\textbf{DAVIS 2017 video object segmentation} performances of the recent self-supervised approaches pre-trained on ImageNet.
The metrics $\mathcal{J}_m$, $\mathcal{F}_m$, and $(\mathcal{J}\& \mathcal{F})_m$ denote mean region similarity, contour-based accuracy, and their average, respectively.
}\label{tbl:davis}
% \vspace{-0.05in}
\end{table}

\begin{table}[t]
\centering
\scalebox{0.95}{
% \resizebox{0.95\linewidth}{!}{
\begin{tabular}{ccccc}
\toprule
% $\mathcal{N}^{(i)}$
                     & Neighbors $\mathcal{N}^{(i)}$ & Matching & Agg       & $(\mathcal{J}\&\mathcal{F})_m$ \\ \midrule
(a)                  & -             & -          & -          & 55.1          \\ \cmidrule(lr){1-5}
\multirow{3}{*}{(b)} & $3\times 3$   & $k=4$      & \checkmark & \textbf{57.0} \\ 
                     & $5\times 5$   & $k=4$      & \checkmark & 56.5          \\
                     & All patches   & $k=4$      & \checkmark & 47.3          \\ \cmidrule(lr){1-5}
\multirow{4}{*}{(c)} & $3\times 3$   & $k=1$      & \checkmark & 56.3          \\
                     & $3\times 3$   & $k=2$      & \checkmark & 56.4          \\
                     & $3\times 3$   & $k=4$      & \checkmark & \textbf{57.0} \\ 
                     & $3\times 3$   & $k=8$      & \checkmark & 56.5          \\ \cmidrule(lr){1-5}
\multirow{2}{*}{(d)} & $3\times 3$   & $k=4$      & -          & 51.4          \\
                     & $3\times 3$   & $k=4$      & \checkmark & \textbf{57.0} \\ 
\bottomrule
\end{tabular}}
\vspace{-0.05in}
\caption{\textbf{Ablation studies on component contributions} of \mbox{SelfPatch}: (a) the baseline (\ie, DINO \cite{caron2021emerging} without SelfPatch), (b) the neighboring patches (``Neighbors $\mathcal{N}^{(i)}$''), (c) positive matching (``Matching''), and (d) aggregation module (``Agg''). All models are pre-trained on COCO \cite{lin2014microsoft} and evaluated on DAVIS 2017~\cite{pont20172017}.}\label{table:contribution}
\vspace{-0.05in}
\end{table}

In this section, we perform ablation studies to understand further how does SelfPatch works. 
To this end, we pre-train all cases
using ViT-Ti/16 on the MS COCO $\tt train2017$ dataset for 200 epochs with a batch size of 256 and evaluate the pre-trained models on the DAVIS 2017 benchmark~\cite{pont20172017}.

\subsection{Component analysis}\label{sec:component}
\cref{table:contribution} shows the individual effects of SelfPatch's components: (b) neighboring patches (``Neighbors'' column), (c) positive matching (``Matching'' column), and (d) the aggregation module (``Agg'' column). Note that (a) shows the performance of the baseline, \ie, DINO \cite{caron2021emerging} without SelfPatch. In what follows, we analyze their effects in detail.

\vspace{0.03in}
\noindent{\bf Effect of neighboring patches.} We investigate the importance of neighboring patches $\mathcal{N}^{(i)}$ for selecting positive patches $\mathcal{P}^{(i)}$ for each patch $\mathbf{x}^{(i)}$. To this end, we test three different types of neighbors: $3\times3$ or $5\times5$ patches around $\mathbf{x}^{(i)}$, or all patches (\eg, $14\times14$ patches in the input resolution of $224\times224$) in the given image $\mathbf{x}$. As shown in the (b) group in \cref{table:contribution}, considering patches far from $\mathbf{x}^{(i)}$ is not helpful for selecting positives $\mathcal{P}^{(i)}$; in particular, considering all patches shows even worse performance than the baseline (\ie, 55.1 $\rightarrow$ 47.3) on the video segmentation task. This validates that restricting positive candidates as neighboring patches is a more reasonable way to find more reliable positive patches than considering far patches.

\vspace{0.03in}
\noindent{\bf Effect of positive matching.} Our positive matching process selects \emph{top-k} patches as positives $\mathcal{P}^{(i)}$ among the neighbors $\mathcal{N}^{(i)}$ using the cosine similarity on the representation space. To demonstrate the effectiveness of this process, we vary the number of positive patches, $k=|\mathcal{P}^{(i)}|$. As shown in the (c) group in \cref{table:contribution}, any $k\in\{1,2,4,8\}$ shows better performance than (a) the baseline. Also, selecting $k=4$ patches among the neighborhood in the positive matching module shows the best performance (\ie, 57.0 $(\mathcal{J}\&\mathcal{F})_m$) compared to selecting all neighboring patches (\ie, $k=8$) and few neighboring patches (\ie, $k\in\{1,2\}$). This result implies that some of the neighboring patches might not be positive (\ie, noisy), but a moderate number of them could be 
considered as positives.

\vspace{0.03in}
\noindent{\bf Effect of aggregation module.} We here validate the contribution of our aggregation module $h_\theta$ which aims at aggregating multiple patch representations. To this end, we simply replace the aggregation module with the average pooling operation. {Note that this pooling operation utilizes patch representations equally while our aggregation module can summarize them in an adaptive manner. Hence, even if a noisy (\eg, not positive) patch exists among the selected positives $\mathcal{P}^{(i)}$, our module $h_\theta$ could reduce the negative effect from the noises. The (d) group in \cref{table:contribution} shows that the aggregation module is a crucial component of our scheme.}

\subsection{Compatibility analysis}\label{section:ablation:compatibility}
We here validate the compatibility of our method with (a) another image-level SSL scheme, MoBY \cite{xie2021self}, and (b) another Transformer-based architecture, Swin Transformer \cite{liu2021swin}. Note that MoBY \cite{xie2021self} is a contrastive method that utilizes negative pairs, and Swin Transformer \cite{liu2021swin} is a variant of ViT, which has a hierarchical structure. \cref{table:moby} summarizes the compatibility experiments.

As shown in \cref{table:moby}, our method is well-incorporated with MoBY \cite{xie2021self}. For example, SelfPatch improves MoBY by 4.8 (\ie, 54.1 $\rightarrow$ 58.9) and 5.6  (\ie, 50.8 $\rightarrow$ 56.4) points when using ViT-Ti/16 and Swin-T architectures, respectively. Since MoBY is a contrastive method, we also construct negative pairs for our loss $\mathcal{L}^{\text{SelfPatch}}$ by sampling global representations 
(\ie, \cref{eq:global_output}) 
of different images. Interestingly, even under such a contrastive 
method,
% scheme,
SelfPatch is not required to use negative pairs: SelfPatch without negatives achieves the competitive performance compared to SelfPatch with the negatives (58.4 \textit{vs.} 58.9). Since the current negative construction for patches is straightforward, we believe that an elaborate design for the negatives might further improve the performance. 
We leave it for future work.

We also found that our method, SelfPatch, is working with a smaller size (\ie, $8\times8$) of input patches as shown in the last two rows in \cref{table:moby}. Note that using such smaller patches is beneficial to solving dense prediction tasks since their final representations are more fine-grained. As a result, DINO + SelfPatch with ViT-Ti/8~\cite{touvron2020training} achieves the best performance in \cref{table:moby}. These results demonstrate the high compatibility of SelfPatch with various SSL schemes, architectures, and patch sizes.

\begin{table}[t]
\centering
\scalebox{0.95}{
% \resizebox{0.95\linewidth}{!}{
\begin{tabular}{lccc}
\toprule
Method & Backbone & Negative & $(\mathcal{J}\&\mathcal{F})_m$ \\
\midrule
MoBY & ViT-Ti/16 & - & 54.1 \\
+ SelfPatch (ours) & ViT-Ti/16 & - & 58.4 \\
+ SelfPatch (ours) & ViT-Ti/16 & \checkmark & \textbf{58.9} \\
\cmidrule(lr){1-4}
MoBY & Swin-T & - & 50.8 \\
+ SelfPatch (ours) & Swin-T & \checkmark & \textbf{56.4}\\
\cmidrule(lr){1-4}
DINO & ViT-Ti/16 & - & 55.1 \\
+ SelfPatch (ours) & ViT-Ti/16 & - & \textbf{57.0}\\
\cmidrule(lr){1-4}
DINO & ViT-Ti/8 & - & 61.6 \\
+ SelfPatch (ours) & ViT-Ti/8 & - & \textbf{65.8}\\
\bottomrule
\end{tabular}}
\vspace{-0.05in}
\caption{\textbf{Ablation studies on compatibility} with various SSL schemes (DINO~\cite{caron2021emerging}, MoBY~\cite{xie2021self}), architectures (ViT \cite{touvron2020training}, Swin Transformer~\cite{liu2021swin}), and patch sizes ($16\times 16$, $8\times 8$).
All models are pre-trained on COCO~\cite{lin2014microsoft} and evaluated on DAVIS 2017~\cite{pont20172017}.
We denote the use of negative pairs for our method as ``Negative''.}
\label{table:moby}
\vspace{-0.05in}
\end{table}

\subsection{Hyperparameter analysis}\label{sec:hyperparameter}
In \cref{table:lossweight}, we examine the effect of the loss weight $\lambda$ in \cref{loss:total} across an array of $\lambda\in\{0,0.1,1\}$.
We observed that the performance of $\lambda=0$ (\emph{i.e.,} 55.3 $(\mathcal{J}\&\mathcal{F})_m$) shows the almost similar to the baseline, DINO (\emph{i.e.,} 55.1 $(\mathcal{J}\&\mathcal{F})_m$).
It shows that the improvement of our method comes from the patch-level self-supervision in \cref{loss:PASS}, not the architectural modification (\emph{i.e.,} adding the aggregation module). As shown in the last two rows in \cref{table:lossweight}, any $\lambda\in\{0.1,1\}$ shows better performance than the baselines. Based on this result, we conducted all experiments in \cref{sec:exp} and \ref{sec:ablation} with $\lambda=0.1$. Remark that this choice of $\lambda=0.1$ shows consistent improvements across architectures (ViT \cite{dosovitskiy2020image, touvron2020training} and Swin Transformer \cite{liu2021swin}), and SSL objectives (DINO \cite{caron2021emerging} and MoBY \cite{xie2021self}) as shown in \cref{section:ablation:compatibility}.

\begin{table}[t]
\centering
\scalebox{0.95}{
% \resizebox{0.95\linewidth}{!}{
\begin{tabular}{lcccc}
\toprule
Method & $\lambda$ & $(\mathcal{J}\&\mathcal{F})_m$ & $\mathcal{J}_m$ & $\mathcal{F}_m$ \\
\midrule
DINO & - & 55.1 & 52.8 & 57.4\\
\cmidrule(lr){1-5}
\multirow{3}{*}{+ SelfPatch (ours)}
 & 0.0 & 55.3 & 53.1 & 57.6\\
 & 0.1 & \textbf{57.0} & \textbf{56.1} & \textbf{57.8}\\
 & 1.0 & 56.4 & 55.5 & 57.4\\
\bottomrule
\end{tabular}
}
\vspace{-0.05in}
\caption{\textbf{Ablation studies on varying the loss weight $\lambda$} for \mbox{SelfPatch} objective. 
All models are pre-trained on COCO \cite{lin2014microsoft} and
evaluated on DAVIS 2017~\cite{pont20172017}.}
\label{table:lossweight}
\vspace{-0.1in}
\end{table}

\section{Discussion}
\noindent{\bf Conclusion.}
We propose a simple yet effective patch-level self-supervised learning scheme (SelfPatch) for improving visual representations of individual image patches. 
Our key idea is to 
treat semantically similar neighboring patches as positives.
More specifically, we
select semantically similar ({\it i.e.}, positive) patches from the neighborhood,
and then 
enforce invariance against each patch and its positives by
minimizing the distance of their final representations.
Through the extensive experiments, we demonstrate the effectiveness of our scheme in various 
downstream tasks, including object detection and semantic segmentation.
We believe that this work would guide many research directions for self-supervised learning.

\vspace{0.03in}
\noindent{\bf Limitation.} {In this work, we focus on constructing \emph{patch-level} self-supervision. However, there might be other types of self-supervision, \eg, more coarse-grained (\ie, region-level) one. Hence, an interesting future research direction would be to develop a universal self-supervision for learning all types (\eg, patch, region, and image-level) of representations. Even for this direction, we believe that our idea of treating neighbors as positives could be utilized.}

\vspace{0.03in}
\noindent{\bf Potential negative societal impact.}
Due to the absence of supervision, self-supervised learning often requires a large number of training samples 
(\eg, DALL-E \cite{ramesh2021dalle} requires 250M image-text pair of data) 
or a large network capacity
(\eg, GPT-3 \cite{brown2020_gpt3} requires 175B parameters).
Such an enormous computational cost may cause environmental 
problems such as carbon generation or global warming. 
Hence, developing various types 
(\eg, pixel- or patch-level) 
of self-supervision for efficient training would be required to alleviate the problems.

\vspace{0.03in}
\noindent{\bf Acknowledgements.}
We thank Sangwoo Mo for
providing helpful feedbacks and suggestions.
This work was supported by Institute of Information \& communications Technology Planning \& Evaluation 
(IITP) grant funded by the Korea government (MSIT) (No.2021-0-02068, Artificial Intelligence 
Innovation Hub; No.2019-0-00075, Artificial Intelligence Graduate School Program (KAIST)) and
the Engineering Research Center Program through the National Research Foundation of Korea (NRF) funded by the Korean Government MSIT (NRF-2018R1A5A1059921).

\newpage

%%%%%%%%% REFERENCES
{\small
\bibliographystyle{ieee_fullname}
\bibliography{egbib}
}

\onecolumn
\clearpage
\begin{center}{\bf {\LARGE Supplementary Material}}
\end{center}
\vspace{0.05in}
\appendix

\section{Pre-training details}\label{app:traning}
For unsupervised pre-training, we use the ImageNet~\cite{deng2009imagenet} dataset for large-scale pre-training (see Sec. 4) and the MS COCO \cite{lin2014microsoft} dataset with the $\tt train2017$ split for medium-scale pre-training (see Sec. 5). Code is available at \url{https://github.com/alinlab/SelfPatch}.

\vspace{0.03in}
\noindent\textbf{ImageNet pre-training details.}
In Sec.~4, we perform unsupervised pre-training using ViT-S/16~\cite{touvron2020training} on the ImageNet~\cite{deng2009imagenet} dataset
for 200 epochs with a batch size of 1024. 
In the case of the joint usage of DINO~\cite{caron2021emerging} and our method (\ie, DINO + SelfPatch), we generally follow the training details of Caron \etal~\cite{caron2021emerging}, including the optimizer and the learning rate schedule. Specifically, we use the AdamW~\cite{loshchilov2018fixing} optimizer with a linear warmup of the learning rate during the first 10 epochs, and the learning rate is decayed with a cosine schedule. We also follow the linear scaling rule~\cite{goyal2017accurate}: $lr = 0.0005 \cdot \text{batchsize}/256$. 
We use 2 global crops and 8 local crops (\ie, $2\times224^2 + 8\times96^2$) for multi-crop augmentation~\cite{caron2020unsupervised, caron2021emerging}.
For our aggregation module, we use two class-attention blocks~\cite{touvron2021going} without Layerscale normalization~\cite{touvron2021going}. 
For the final output dimension of the projection head, we use $K=65536$ for the SSL projection head following Caron \etal~\cite{caron2021emerging} and $K=4096$ for our projection head.
In Sec. 4, we use a publicly available DINO pre-trained model\footnote{\url{https://github.com/facebookresearch/vissl}.} with 300 training epochs on the ImageNet as the baseline, which also use the same hyperparameters as the above.

\vspace{0.03in}
\noindent\textbf{MS COCO pre-training details.} In Sec. 5,
we perform unsupervised pre-training using ViT-Ti/16~\cite{touvron2020training} on the MS COCO \cite{lin2014microsoft} dataset
with $\tt train2017$ split for 200 training epochs with a batch size of 256. 
In the case of the joint usage of DINO~\cite{caron2021emerging} and our method (\ie, DINO + SelfPatch), we use 2 global crops and 2 local crops (\ie, $2\times224^2 + 2\times96^2$) for the multi-crop augmentation and $K=4096$ for both SSL and SelfPatch projection head.
In the case of the joint usage of MoBY~\cite{xie2021self} and our method (\ie, MoBY + SelfPatch), we also perform the pre-training for 200 training epochs with a batch size of 256, and 
follow the training details of Xie \etal~\cite{xie2021self} for both ViT-Ti/16~\cite{touvron2020training} and Swin-T \cite{liu2021swin}.

\section{Evaluation details}\label{app:evaluation}
For evaluation, we perform object detection and instance segmentation on MS COCO~\cite{lin2014microsoft}, semantic segmentation on ADE20K~\cite{zhou2017scene}, and video object segmentation on DAVIS-2017~\cite{pont20172017}. 

\vspace{0.03in}
\noindent\textbf{COCO object detection and instance segmentation.} MS COCO~\cite{lin2014microsoft} is large-scale object detection, segmentation, and captioning dataset: in particular, $\tt train2017$ and $\tt val2017$ splits contain 118K and 5K images, respectively. We follow the basic configuration of $\tt {mmdetection}$\footnote{\url{https://github.com/open-mmlab/mmdetection}.}~\cite{mmdetection} 
for fine-tuning Mask R-CNN~\cite{he2017mask} with FPN~\cite{lin2017feature} under the standard $\tt 1x$ schedule.
In addition, we follow several details of El-Nouby \etal~\cite{el2021xcit} for integrating Mask R-CNN with ViT-S/16.

\vspace{0.03in}
\noindent\textbf{ADE20K semantic segmentation.} ADE20K~\cite{zhou2017scene} is a semantic segmentation benchmark containing 150 fine-grained semantic categories and 25k images. We follow all the configurations of $\tt {mmsegmentation}$\footnote{\url{https://github.com/open-mmlab/mmsegmentation}.}~\cite{mmseg2020}
for fine-tuning Semantic FPN~\cite{kirillov2019panoptic} with 40k iterations and an input resolution of 512×512. 
We also perform large-scale fine-tuning experiments using UPerNet~\cite{xiao2018unified} with 160k iterations and an input resolution of 512×512 in \cref{ade20k:upernet}.

\vspace{0.03in}
\noindent\textbf{DAVIS 2017 video object segmentation.} DAVIS 2017~\cite{pont20172017} is a video object segmentation dataset containing 60 training, 30 validation, and 60 testing videos. We 
follow the evaluation protocol of Jabri~\cite{jabri2020space} and Caron \etal~\cite{caron2021emerging}, which evaluates the quality of frozen representations of
image patches by segmenting scenes with the nearest neighbor between consecutive frames.

\section{UPerNet on ADE20K semantic segmentation}\label{ade20k:upernet}
Here, we additionally evaluate semantic segmentation performances of DINO and DINO + SelfPatch 
for a large-scale fine-tuning setup, \ie, a larger network and longer iterations.
Specifically, we use UPerNet~\cite{xiao2018unified} with 160k iterations following Liu \etal~\cite{liu2021swin}, while Wang~\cite{wang2021pyramid} and El-Nouby \etal~\cite{el2021xcit} do use Semantic FPN~\cite{kirillov2019panoptic} with 40k iterations as we follow originally. 
\cref{tbl:ade:upernet} summarizes the results.
We emphasize that DINO + SelfPatch still achieves consistent improvements over DINO in all the metrics; \eg,
DINO + SelfPatch achieves 0.9, 1.1, and 1.2 points higher than DINO in terms of the mIoU, aAcc, and mAcc metrics, respectively.
This comparison under the large-scale fine-tuning setup also verifies the effectiveness of SelfPatch.

\begin{table}[h]
\centering
\scalebox{0.95}{
\begin{tabular}{llccccc}
\toprule
Method & Network & Param.(M) & Iteration & mIoU & aAcc & mAcc \\
\midrule
DINO \cite{caron2021emerging}& ViT-S/16 + Semantic FPN & 26 & 40k & 38.3 & 79.0 & 49.4\\
+ SelfPatch (ours) & ViT-S/16 + Semantic FPN & 26 & 40k & \textbf{41.2} & \textbf{80.7} & \textbf{52.1}\\ \midrule
DINO \cite{caron2021emerging}& ViT-S/16 + UPerNet & 58 & 160k & 42.3 & 80.4 & 52.7\\
+ SelfPatch (ours) & ViT-S/16 + UPerNet & 58 & 160k & \textbf{43.2} & \textbf{81.5} & \textbf{53.9}\\
\bottomrule
\end{tabular}}
\vspace{-0.05in}
\caption{\textbf{Transferring performances to ADE20K semantic segmentation} using Semantic FPN~\cite{kirillov2019panoptic} and
UPerNet~\cite{xiao2018unified} with 40k and 160k iterations, respectively. All models are pre-trained on the ImageNet~\cite{deng2009imagenet} dataset using ViT-S/16.
The metrics, mIoU, aAcc, and mAcc, denote the mean intersection of union, all pixel accuracy, and mean class accuracy, respectively.
}\label{tbl:ade:upernet}
\vspace{-0.05in}
\end{table}

\section{Linear classification on ImageNet}\label{app:classification}
We here evaluate the quality of pre-trained representations for the image classification task under the conventional linear evaluation protocol~\cite{wu2018unsupervised, chen2020simple, caron2021emerging}. Specifically, we train a supervised linear classifier on top of frozen features without the projection head following the details of Caron \etal~\cite{caron2021emerging}; we use the SGD optimizer with a batch size of 1024 during 100 training epochs and report central-crop top-1 accuracy. \cref{table:classification} summarizes the results.
Here, we would like to emphasize that our primary applications of interest are
dense prediction tasks (i.e., not classification tasks), where patch-level representation learning can be more effective.
Nevertheless, DINO + SelfPatch can outperform DINO even for ImageNet classification under the same 300 training epochs; ours and DINO achieve $75.6\%$ and $75.1\%$, respectively. Also, DINO + SelfPatch consistently outperforms other self-supervised ViT baselines: MoCo-v3~\cite{chen2021empirical} and MoBY~\cite{xie2021self}.
It shows that our method is not only able to enhance the performances on dense prediction tasks, but also maintain competitive performance on image classification.

\begin{table}[h]
\centering
\begin{tabular}{lccc}
\toprule
Method & Backbone & Epoch & Top-1 acc. \\
\midrule
MoCo-v3~\cite{chen2021empirical} & ViT-S/16 & 300 & 73.2\\
MoBY~\cite{xie2021self} & ViT-S/16 & 300 & 72.8\\
\cmidrule(lr){1-4}
DINO~\cite{caron2021emerging} & ViT-S/16 & 300 & 75.1\\
+ SelfPatch (ours) & ViT-S/16 & 300 & {\bf 75.6}\\
\bottomrule
\end{tabular}   
\vspace{-0.05in}
\caption{\textbf{ImageNet linear classification} performances of the recent self-supervised ViTs pre-trained on
the ImageNet~\cite{deng2009imagenet} benchmark. We train a supervised linear classifier on top of frozen features and report central-crop top-1 accuracy.}
\label{table:classification}
\vspace{-0.05in}
\end{table}

\section{Comparison with concurrent work}
Concurrent to our work, EsViT~\cite{li2021esvit} introduces the region-matching (\ie, matching image patches) task for Vision Transformers that 
considers the region correspondence (\ie, matching the two most similar regions) between 
two differently augmented images. In particular, the region-matching task also has been investigated for ResNet; for example, DenseCL~\cite{wang2021dense} also matches the two most similar spatial representations between the two augmented images.
One key
difference from our method
is that
the region-matching task finds positive pairs from two strongly augmented images, which necessarily requires overlapping regions and may find noisy positives (\ie, not positives) in early training, while our method utilizes adjacent patches in the same augmented image as the positives, which is a reasonable way to find reliable positives without constraints of overlapping regions.

To further compare our method with the region matching task, we pre-train EsViT using ViT-Ti/16 on the MS COCO dataset \cite{lin2014microsoft}
(\ie, the same training details in \cref{app:traning}), and perform three evaluation downstream tasks: (a) COCO object detection and instance segmentation, (b) ADE20K semantic segmentation, and (c) DAVIS 2017 video object segmentation.
As shown in \cref{table:esvit}, our method consistently outperforms EsViT with a large
margin in all the metrics, \eg, (a) +2.8 AP$^{\text{bb}}$ on COCO detection, +1.7 AP$^{\text{mk}}$ on COCO detection, (b) +3.5 mIoU on ADE20K segmentation, and (c) +3.5 $(\mathcal{J}\&\mathcal{F})_m$ on DAVIS segmentation.
We believe that restricting positive candidates to neighboring patches plays an essential role in 
constructing effective patch-level self-supervision,
and this work would guide a new research direction for patch-level self-supervised learning.

\begin{table}[h]
\centering
% \scalebox{0.95}{
\resizebox{\textwidth}{!}{
\begin{tabular}{lccccccc|ccc|ccc}
\toprule
& & \multicolumn{3}{c}{COCO Detection} & \multicolumn{3}{c}{COCO Segmentation} & \multicolumn{3}{c}{ADE20K Segmentation} & \multicolumn{3}{c}{DAVIS Segmentation} \\
\cmidrule(lr){3-5}\cmidrule(lr){6-8}\cmidrule(lr){9-11}\cmidrule(lr){12-14}
Method & Backbone & AP$^{\text{bb}}$ & AP$_{50}^{\text{bb}}$ & AP$_{75}^{\text{bb}}$ & AP$^{\text{mk}}$ & AP$_{50}^{\text{mk}}$ & AP$_{75}^{\text{mk}}$ & mIoU & aAcc & mAcc & $(\mathcal{J}\&\mathcal{F})_m$ & $\mathcal{J}_m$ & $\mathcal{F}_m$\\
\midrule
DINO \cite{caron2021emerging}& ViT-Ti/16 & 28.0 & 48.8 & 28.4 & 26.9 & 45.8 & 27.7 & 24.9 & 73.4 & 33.3 & 55.1 & 52.8 & 57.4\\ 
+ SelfPatch (ours) & ViT-Ti/16 & \textbf{30.7} & \textbf{51.4} & \textbf{32.2} & \textbf{28.6} & \textbf{48.2} & \textbf{29.6} & \textbf{29.5} & \textbf{75.5} & \textbf{39.2} & \textbf{57.0} & \textbf{56.1} & \textbf{57.8}\\
\cmidrule(lr){1-14}
EsViT \cite{li2021esvit} & ViT-Ti/16 & 27.9 & 49.0 & 28.0 & 26.9 & 45.9 & 27.7 & 26.0 & 73.5 & 34.5 & 53.5 & 50.8 & 56.2\\
\bottomrule
\end{tabular}
}
\vspace{-0.05in}
\caption{\textbf{Transferring performances to various downstream tasks}: COCO object detection and instance segmentation, ADE20K semantic segmentation, and DAVIS 2017 video object segmentation. All models are pre-trained on the MS COCO~\cite{lin2014microsoft} dataset with $\tt train2017$ split using ViT-Ti/16. We use the same evaluation details in \cref{app:evaluation}.
}\label{table:esvit}
\vspace{-0.05in}
\end{table}

\section{Importance of positional encoding}
In this section, we investigate the importance of positional encoding (PE) in a dense prediction task, similar to Chen~\etal~\cite{chen2021empirical}. Specifically, we pre-train ViT-S/16 models on COCO with or without PE, and evaluate their segmentation performances on DAVIS. Table \ref{table:pe} shows that
learning PE is still effective even under SelfPatch, %for dense prediction tasks, 
while
SelfPatch consistently improves the performance regardless of PE.
It 
implies
that the role of positional inductive bias in a dense prediction task would be quite important, and 
our method, SelfPatch, orthogonally contributes to improving patch-level representations.

\begin{table}[h]
\centering
\scalebox{0.9}{%
\begin{tabular}{lcccc}
\toprule
Method & PE & $(\mathcal{J}\&\mathcal{F})_m$ & $\mathcal{J}_m$ & $\mathcal{F}_m$ \\
\midrule
DINO                    & \checkmark & 55.1 & 52.8 & 57.4 \\
DINO + SelfPatch & \checkmark & \textbf{57.0} & \textbf{56.1} & \textbf{57.8} \\
\midrule
DINO                    &            & 51.7 & 49.5 & 54.0 \\
DINO + SelfPatch &            & \textbf{52.9} & \textbf{50.5} & \textbf{55.2} \\
\bottomrule
\end{tabular}}
\vspace{-0.05in}
\caption{\textbf{Importance of positional encoding (PE).} All models are pretrained on the MS COCO \cite{lin2014microsoft} dataset with $\tt train2017$ split using ViT-S/16. We use the same evaluation details for DAVIS 2017 video object segmentation in \cref{app:evaluation}.}\vspace{-0.05in}
\label{table:pe}
\end{table}

\section{Effects of the number of positive patches under varying patch sizes}
We primarily focus on the popular setup of $224\times224$ images and $16\times16$ patches,
where 
$k=4$ works as we validated 
throughout the paper. 
However, this choice may not be optimal for other setups; 
we additionally perform an ablation study on a different dataset, ImageNet-10~\cite{li2021contrastive},
with
$8\times8$, $16\times16$ and $32\times32$ 
patches from $224\times224$ images.
Table \ref{table:im10} shows their segmentation performances on DAVIS and $20$-NN (\emph{i.e.,} 20
nearest neighbor classifier) classification performances 
following Caron~\etal~\cite{caron2021emerging}.
Overall, it suggests that the effective number of positives may depend on the relative size of patches in an image. 
For example, $k=4,6$ achieves the best performance for $8\times8$ and $16\times16$ patch sizes on both the dense prediction and the classification tasks, while $k=2$ does for the patch size $32\times32$. This is because smaller patches would contain more positive patches in their neighbors. Hence, we recommend to use $k=4$ in general cases (\eg, $8\times8$ and $16\times16$), but $k=2$ when considering a larger patch size (\eg, $32\times32$) for $224\times224$ images.

\begin{table}[h]
\centering
\scalebox{0.9}{%
\begin{tabular}{lc|ccc|ccc}
\toprule
\multicolumn{2}{c|}{Patch size} & \multicolumn{3}{c|}{$(\mathcal{J}\&\mathcal{F})_m$} & \multicolumn{3}{c}{Acc.} \\
Method & $k$ & 32 $\times$ 32 & 16 $\times$ 16 & 8 $\times$ 8 & 32 $\times$ 32 & 16 $\times$ 16 & 8 $\times$ 8\\\midrule
DINO       & - & 24.9 & 37.6 & 48.7 & 76.0 & 83.8 & 85.0\\ \midrule
\multirow{5}{*}{DINO + SelfPatch}
      & 1 & 34.9 & 46.5 & 50.6 & 80.0 & 85.0 & 87.0\\
      & 2 & \textbf{36.5} & 52.9 & 57.3 & \textbf{80.2}& \textbf{86.0} & 88.0\\
      & 4 & 36.3 & \textbf{53.1} & 61.7 & 75.0 & 85.8 & \textbf{88.4}\\
      & 6 & 36.3 & 52.6 & \textbf{62.0} & 75.6 & 85.0 & 87.4\\
      & 8 & 33.2 & 50.4 & 60.4 & 75.2 & 82.6 & 87.2\\
\bottomrule
\end{tabular}}
\vspace{-0.05in}
\caption{Effects of the positive number $k$ under varying the different patch sizes. All models are pre-trained on ImageNet-10~\cite{li2021contrastive}, and evaluated on DVAIS video segmentation and ImageNet-10 classification.} \vspace{-0.05in}
\label{table:im10}
\end{table}

In addition,
we count the number of positive patches in the COCO~\cite{lin2014microsoft} validation images by using their ground-truth segmentation labels
%on the COCO~\cite{lin2014microsoft} validation data by using their ground-truth segmentation maps
and found $4.3{\pm}1.2\;(k\approx4)$ adjacent positives (on average) for $16\times16$ patches from $224\times224$ images. 
Here, we measure the cosine similarities among adjacent patches and use the threshold of 0.95 for counting the positives.
Interestingly, we observe that further utilizing the ground-truth segmentation labels for the adjacent positive selection can improve ours from 57.0 to 59.5 $(\mathcal{J}\&\mathcal{F})_m$ score on DAVIS~\cite{pont20172017}.
We believe that developing an unsupervised adaptive selection scheme on $k$ would be an interesting direction to explore.

\end{document}